\documentclass{article}
\usepackage{spconf,amsmath,graphicx}

\usepackage{enumitem}
\setlist{nosep, leftmargin=14pt}

\usepackage{mwe} % to get dummy images
\usepackage{graphicx}
\usepackage{booktabs}
\usepackage{amsmath}
\usepackage{amsfonts}
\usepackage{multirow}
\usepackage{xcolor}
\usepackage{lipsum}
\usepackage{marvosym}
\usepackage{enumitem}
\usepackage{url}
\usepackage{breakurl}
\title{Center-Aware Detection with Swin-based Co-DETR Framework for Cervical Cytology}
\name{Yan Kong$^{1}$,  Yuan Yin$^{2}$, Hongan Chen$^{1}$, Yuqi Fang $^{1,3 *}$, Caifeng Shan$^{1,3 *}$ \thanks{$*$ Corresponding author.}
}
\address{
$^{1}$ School of Intelligence Science and Technology, Nanjing University, Nanjing, China. \\
$^{2}$ School of Biomedical Engineering, ShanghaiTech University, Shanghai, China. \\
$^{3}$ State Key Laboratory for Novel Software Technology, Nanjing University, Nanjing, China.}
\begin{document}
%\ninept
%
\maketitle

\begin{abstract}

Automated analysis of Pap smear images is critical for cervical cancer screening but remains challenging due to dense cell distribution and complex morphology. 
In this paper, we present our winning solution for the RIVA Cervical Cytology Challenge, achieving 1st place in Track B and 2nd place in Track A. 
Our approach leverages a powerful baseline, integrating the Co-DINO framework with a Swin-Large backbone for robust multi-scale feature extraction. 
To address the dataset's unique fixed-size bounding box annotations, we formulate the detection task as a center-point prediction problem. 
Tailoring our approach to this formulation, we introduce a center-preserving data augmentation strategy and an analytical geometric box optimization to effectively absorb localization jitter. 
Finally, we apply track-specific loss tuning to adapt the loss weights for each task. 
Experiments demonstrate that our targeted optimizations improve detection performance, providing an effective pipeline for cytology image analysis. 
Our code is available at \url{https://github.com/YanKong0408/Center-DETR}.
\end{abstract}

\section{Introduction}

Cervical cancer screening via Papanicolaou (Pap) smears is a vital clinical procedure. However, manual slide inspection is labor-intensive, time-consuming, and subject to inter-observer variability. This underscores the urgent need for automated, high-precision analysis systems.

To advance this critical domain, the RIVA Cervical Cytology Challenge~\cite{riva-cervical-cytology-challenge-track-b-isbi-final-evaluation} provides an exceptionally well-curated and large-scale dataset~\cite{perez2025riva}. By offering rigorous independent annotations, this challenge serves as a highly valuable benchmark for the medical imaging community. It evaluates algorithms across two distinct tracks: Track A for localization and classification (into eight Bethesda categories), and Track B strictly for cell localization.

Despite the availability of modern detection architectures, applying them directly to the RIVA dataset reveals several unique bottlenecks: (1) \textbf{Small and Dense Objects:} Cells occupy a tiny fraction of high-resolution images and are often densely clustered. (2) \textbf{Annotation Constraints and Size Bias:} We observe a critical annotation characteristic where ground-truth bounding boxes are rigidly fixed at $100 \times 100$ pixels. Standard regression-based detectors struggle with this, tending to predict excessively large boxes to mitigate the Intersection over Union (IoU) penalty caused by minor center-point shifts. This introduces severe size regression noise and degrades evaluation metrics.
(3) \textbf{Morphological Integrity:} Accurate classification relies heavily on intact cellular structures (e.g., nucleocytoplasmic ratio), which are easily destroyed by standard random cropping.

Motivated by these insights, we reformulate the task as a center-aware detection problem. Our proposed framework, which achieved 1st place in Track B and 2nd place in Track A, makes the following contributions:
\begin{itemize}
    \item We establish a powerful baseline utilizing the Co-DINO~\cite{zong2023detrs} framework with a Swin-Large backbone~\cite{liu2021swin}, exploiting collaborative auxiliary supervision and hierarchical multi-scale features to localize dense, small cells.
    \item We propose a Center-Preserving Data Augmentation strategy that discards severely truncated cells, ensuring the network learns strictly from biologically valid morphology.
    \item We introduce an Analytical Geometric Box Optimization to reconstruct fixed-size ($101.5 \times 101.5$) bounding boxes from predicted centers. This model-agnostic technique explicitly mitigates localization jitter.
    \item We apply Track-Specific Loss Tuning to dynamically adapt the network's optimization focus to the distinct evaluation criteria of each track.
\end{itemize}

\begin{figure*}[ht]
    \centering
    \includegraphics[width=1\linewidth]{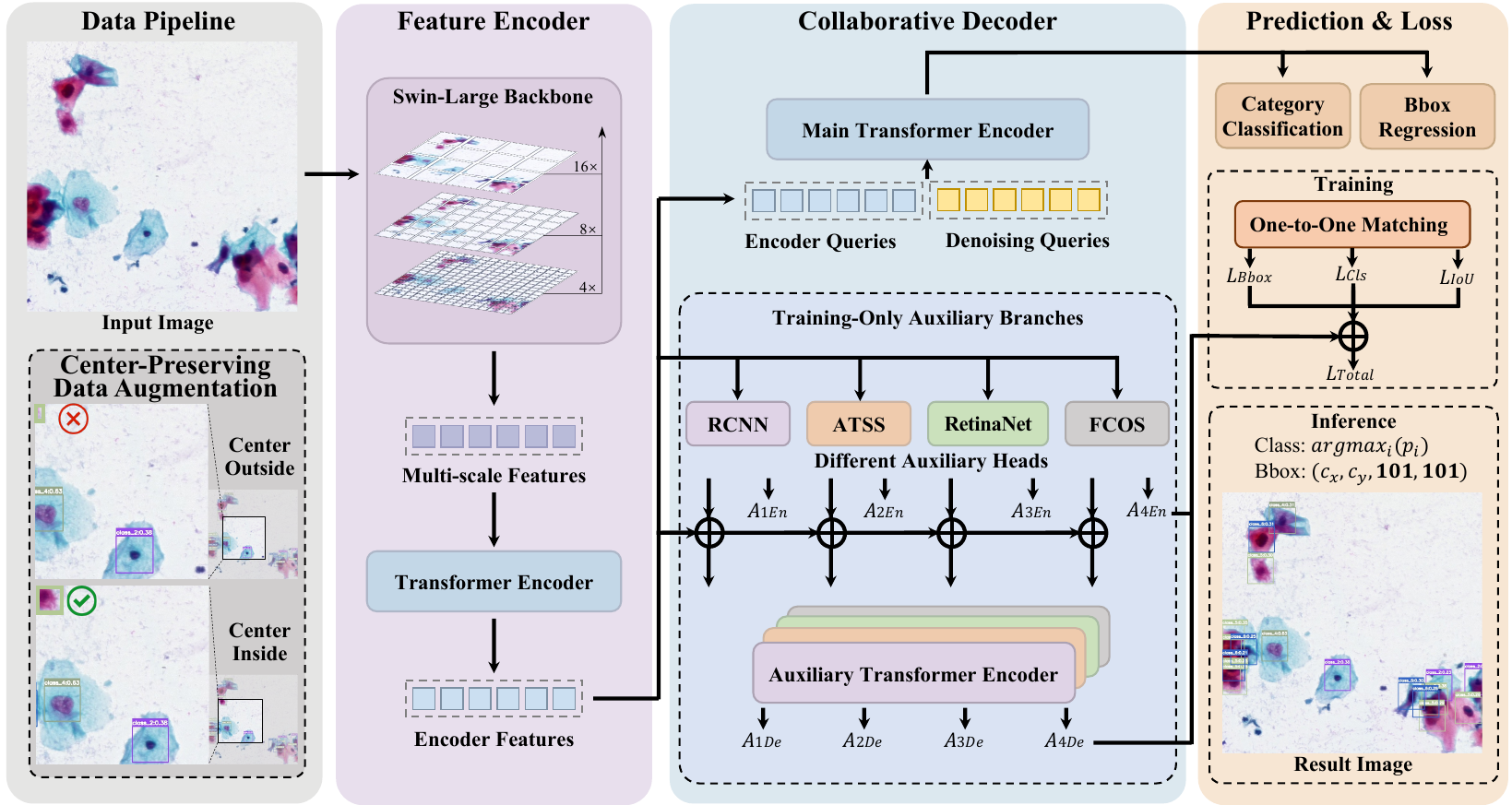}
    \caption{\textbf{Overview of the proposed center-aware Co-DINO framework.}}
    \label{fig:framework}
\end{figure*}

\section{Methodology}

In this work, we formulate the detection problem as a center-based detection task, in which the model predicts object centers and their associated categories. 
Given an input image $I$, the detector predicts a set of object hypotheses
\[
\hat{\mathcal{B}} = \left\{ \left( \hat{c}_{x_i}, \hat{c}_{y_i}, \hat{\mathbf{p}}_i \right) \right\}_{i=1}^{N},
\]
where $(\hat{c}_{x_i}, \hat{c}_{y_i})$ denotes the predicted center coordinates of the $i$-th object, and 
$\hat{\mathbf{p}}_i \in \mathbb{R}^{9}$ represents the predicted class probability vector.
The 9-dimensional probability vector corresponds to the eight predefined Bethesda categories and one additional background class. 

Our solution is as illustrated in Figure \ref{fig:framework}. The detailed components are described in the following subsections.

\subsection{Co-DINO Framework with Swin-Large Backbone}

Our detection pipeline adopts Co-DINO with a Swin-Large backbone, combining hierarchical feature extraction and collaborative hybrid assignment to handle the dense distribution and subtle inter-class variations in cervical cytology images.

\textbf{Hierarchical Feature Extraction:} While state-of-the-art detectors on natural image benchmarks predominantly employ ViT-Large with fixed-scale patch processing, we intentionally select Swin-Large for cytology analysis. Its shifted localized attention windows progressively build hierarchical multi-scale representations. This denser patch modeling scheme grants the network superior sensitivity to small spatial details, making it well-suited for localizing tightly clustered cells in Pap smears.

\textbf{Collaborative Decoding and Auxiliary Supervision:} Standard DETR-like models rely on strictly one-to-one bipartite matching, which often leads to sparse supervision and limits the discriminative capability of the encoder. To overcome this, Co-DINO introduces a Training-Only Auxiliary Branch module that employs multiple classic detection heads in parallel. Each head operates on one-to-many label assignments to provide dense, diverse supervision:

\textbf{Faster R-CNN:} A two-stage proposal-based head that refines for high-precision localization.

\textbf{ATSS:} An adaptive anchor matching head bridging anchor-based and anchor-free learning dynamically.

\textbf{RetinaNet:} A classic anchor-based head providing dense, scale-aware feature supervision.

\textbf{FCOS:} An anchor-free, center-based predictor which perfectly aligns with our center-point task formulation.

These auxiliary heads enrich the positive supervision to assist the sparse queries of the main decoder. Furthermore, denoising queries~\cite{zhangdino} are injected during training to optimize a reconstruction objective. This enforces instance-level consistency under spatial perturbations, substantially improving the model's robustness to localization errors in crowded regions. All auxiliary branches are discarded during inference, yielding a highly accurate yet efficient detector.

\subsection{Center-Preserving Data Augmentation}

Standard random cropping retains all bounding boxes partially overlapping the cropped region. However, this is unsuitable for cervical cytology, where diagnostic cues depend on nucleus morphology and the nucleus-to-cytoplasm ratio. If the nucleus center lies outside the crop, the visible fragment lacks meaningful diagnostic information and introduces noisy supervision.

To address this issue, we adopt a center-preserving cropping strategy: given ground-truth center coordinates $(c_{x_i}, c_{y_i})$ and a cropped region $R = [x_{\min}, x_{\max}] \times [y_{\min}, y_{\max}]$, an object is retained only if its center satisfies $(c_{x_i}, c_{y_i}) \in R$, while objects with centers outside $R$ are discarded.

This ensures that retained objects preserve valid nucleus morphology and eliminates misleading supervision from incomplete cells. It also improves robustness to partial cell observations while maintaining reliable center-based supervision.

\subsection{Analytical Geometric Bounding Box Optimization}
\label{section2.3}
Since we formulate detection as a center-point prediction task, to optimize the bounding box dimensions for maximizing $mAP$, we directly post-process all predictions to a fixed size of $w = h = 101.5$. The rationale is as follows.

Given a ground-truth box $\mathcal{B}_{gt}$ with side length $w_{gt} = h_{gt} = 100$, and a predicted box $\mathcal{B}_{pred}$ with center $(x_c, y_c)$ and unified side length $S$, the intersection sizes under center localization errors $\Delta_x = |x_c - x_{gt}|$ and $\Delta_y = |y_c - y_{gt}|$ are:
\[
W_{int} = \min\left(100, \max\left(0, \frac{100 + S}{2} - \Delta_x\right)\right),
\]
\[
H_{int} = \min\left(100, \max\left(0, \frac{100 + S}{2} - \Delta_y\right)\right).
\]

Setting $S = 100$ causes any center shift to strictly penalize IoU. By expanding $S > 100$, we introduce a spatial buffer $\gamma = (S - 100) / 2$ that preserves intersection area when $\Delta \leq \gamma$. The optimal size $S^*$ maximizes the expected IoU over the empirical distribution of localization jitter $\epsilon \sim (\Delta_x, \Delta_y)$:
\[
S^* = \arg \max_{S} \mathbb{E}_{\epsilon} \left[ \frac{W_{int} \cdot H_{int}}{100^2 + S^2 - (W_{int} \cdot H_{int})} \right].
\]

With typical jitter of $\epsilon \approx 1 \sim 1.5$ pixels, slightly expanding the box marginally increases the union area while significantly preserving intersection. Our derivation confirms that $w = h = 101.5$ yields the optimal expected IoU, directly maximizing the $mAP$ metric.

\subsection{Track-Specific Loss Reweighting}

To train the collaborative architecture, the overall loss function incorporates both the main decoder loss and the auxiliary supervision losses. The total loss $\mathcal{L}_{Total}$ is formulated as:
\[
\mathcal{L}_{Total} = \lambda_{cls}\mathcal{L}_{focal} + \lambda_{L1}\mathcal{L}_{L1} + \lambda_{IoU}\mathcal{L}_{GIoU} + \lambda_{Aux}\mathcal{L}_{Aux},
\]
where $\mathcal{L}_{Aux}$ aggregates losses from auxiliary heads ($\mathcal{A}_{1En}$ to $\mathcal{A}_{4En}$ and $\mathcal{A}_{1De}$ to $\mathcal{A}_{4De}$) via one-to-many assignments. For the main one-to-one branch, $\mathcal{L}_{focal}$ governs classification, while $\mathcal{L}_{L1}$ and $\mathcal{L}_{GIoU}$ supervise box regression.

As detailed in Section~\ref{section2.3}, our geometric post-processing fixes bounding box dimensions to $101.5 \times 101.5$, mathematically optimizing the expected $mAP$. This geometric prior substantially reduces the need for network-driven size regression, allowing us to aggressively down-weight the GIoU loss ($\lambda_{IoU} = 0.5$) and adaptively tune the remaining weights to address each track's distinct bottleneck:

\begin{itemize}
    \item \textbf{Track A (Detection + Classification):} Fine-grained discrimination among visually similar Bethesda categories is the primary challenge. We prioritize classification by setting $\lambda_{cls} = 3.5$ and $\lambda_{L1} = 1.5$.
    \item \textbf{Track B (Detection Only):} With no class evaluation, $mAP$ depends solely on center-point precision. We prioritize localization by setting $\lambda_{L1} = 3.5$ and $\lambda_{cls} = 1.5$.
\end{itemize}

This decoupled reweighting aligns gradient updates with each track's evaluation constraints, while the dense $\mathcal{L}_{Aux}$ stabilizes convergence of the deep Transformer encoder.

\section{Experiments and Results}
%\section{Experiments and Results}

\subsection{Dataset and Implementation Details}

We conducted experiments on the official RIVA challenge dataset~\cite{perez2025riva}. For experiments, models were trained on the training split and evaluated on the validation split. For the challenge submission, we merged both splits to maximize training data utilization.

Our implementation is based on MMDetection~\cite{chen2019mmdetection} with PyTorch. Models were initialized with COCO pre-trained weights and trained using default hyperparameters. Training was conducted on an NVIDIA RTX 4090 (48GB VRAM); inference requires only 12GB VRAM, indicating practical feasibility for clinical deployment.

\begin{table}[h]
\caption{Performance Comparison on RIVA Challenge. Bold and underline indicate the best and second-best results.}
\label{tab:results}
\centering
\begin{tabular}{l|cc}
\hline
Model              & Track A (mAP)  & Track B (mAP)  \\ \hline
YOLO~\cite{redmon2018yolov3}             & 0.120          & 0.460          \\
RetinaNet~\cite{lin2017focal}          & 0.137          & 0.507          \\
CenterNet~\cite{duan2019centernet}          & 0.063          & 0.519          \\
Co-Deformable-DETR & 0.207          & 0.586          \\
Co-DINO-ViT        & \textbf{0.238} & \underline{0.604} \\
Co-DINO-Swin (Ours)& \underline{0.237} & \textbf{0.609} \\ \hline
\end{tabular}
\end{table}

% \subsection{Performance Comparison}
% Table \ref{tab:results} summarizes the performance of our method against various baselines.The results indicate that Transformer-based detectors significantly outperform traditional CNN-based models. Specifically, our Swin-Large backbone demonstrates superior localization precision in Track B, while maintaining competitive classification performance in Track A.

% \subsection{Ablation Study}
% To evaluate the contribution of each component, we conducted ablation studies as shown in Table \ref{tab:ablation}.

% \begin{table}[h]
% \caption{Ablation Study of Proposed Refinements}
% \label{tab:ablation}
% \centering
% \begin{tabular}{ccc|cc}
% \hline
% C. C. & B. O.  & L. T. & mAP(Track B) & mAP(Track A) \\ \hline
%             &                  &             & 0.609 & 0.604 \\
% \checkmark  &                  &             & 0.611 & 0.606 \\
% \checkmark  & \checkmark       &             & 0.634 & 0.620 \\
% \checkmark  & \checkmark       & \checkmark  & \textbf{0.635}  & \textbf{0.628}\\ \hline
% \end{tabular}
% \end{table}

% The integration of center-preserving cropping ensures morphological integrity, and the $101.5 \times 101.5$ box optimization effectively absorbs localization jitter. Finally, track-specific loss reweighting aligns the model with the distinct evaluation metrics of each track.

\subsection{Performance Comparison}

Table~\ref{tab:results} compares our method against various baselines. Transformer-based detectors significantly outperform CNN-based models. Our Swin-Large backbone achieves superior localization in Track B while maintaining competitive classification in Track A.

\subsection{Ablation Study}

Table~\ref{tab:ablation} quantifies the contribution of each component. Center-preserving cropping (C.C.) improves morphological integrity, box optimization (B.O.) absorbs localization jitter, and track-specific loss tuning (L.T.) aligns optimization with track-specific metrics.

\begin{table}[h]
\caption{Ablation Study in Track B (mAP)}
\label{tab:ablation}
\centering
\begin{tabular}{l|cc}
\hline
Component & Co-DINO-Swin & Co-DINO-ViT \\ \hline
Baseline & 0.609 & 0.604 \\
+ C.C. & 0.611 & 0.606 \\
+ C.C. + B.O. & 0.634 & 0.620 \\
+ C.C. + B.O. + L.T. & \textbf{0.635} & \textbf{0.628} \\ \hline
\end{tabular}
\end{table}

% \subsection{Effectiveness of Geometric Box Optimization}

% To further validate the robustness of the proposed center-based box optimization ($w=h=101.5$), we applied this post-processing technique to all baseline models in Track B. As shown in Table \ref{tab:baseline_boost}, every evaluated architecture experienced a significant performance gain.

% \begin{table}[h]
% \caption{Performance gain of baseline models in Track B with the proposed $101.5 \times 101.5$ box optimization.}
% \label{tab:baseline_boost}
% \centering
% \begin{tabular}{l|cc}
% \hline
% Model              & Baseline mAP & With Box Opt. \\ \hline
% YOLO               & 0.460        & 0.478          \\
% RetinaNet          & 0.507        & 0.523          \\
% CenterNet          & 0.519        & 0.525          \\
% Co-Deformable-DETR & 0.586        & 0.601          \\
% Co-DINO-ViT       & 0.604        & 0.618 \\ \hline
% \end{tabular}
% \end{table}

% This universal improvement reveals a critical optimization bias: under standard IoU loss, models inherently tend to predict larger boxes to mitigate the steep IoU penalty caused by center-point displacement. By analytically fixing the dimensions to $101.5 \times 101.5$ during post-processing, we effectively absorb this localization jitter and bypass the noisy size regression. These results confirm that our geometric prior is a model-agnostic refinement, particularly valuable for any detection task characterized by fixed-size annotations.

\subsection{Effectiveness of Geometric Box Optimization}

Given its substantial and universal impact, we dedicate a separate analysis to the $101.5 \times 101.5$ box optimization. As shown in Table \ref{tab:baseline_boost}, this simple post-processing step consistently boosts performance across all baseline architectures in Track B. The consistent gains confirm that this geometric prior effectively absorbs localization jitter without modifying model architecture—making it a highly practical enhancement for detection tasks with fixed-size annotations.

\begin{table}[h]
\caption{Performance gain of baseline models in Track B with the proposed $101.5 \times 101.5$ box optimization.}
\label{tab:baseline_boost}
\centering
\begin{tabular}{l|cc}
\hline
Model              & Baseline mAP & With Box Opt. \\ \hline
YOLO               & 0.460        & 0.478 $\uparrow$         \\
RetinaNet          & 0.507        & 0.523 $\uparrow$         \\
CenterNet          & 0.519        & 0.525 $\uparrow$         \\
Co-Deformable-DETR & 0.586        & 0.601 $\uparrow$          \\
Co-DINO-ViT       & 0.604        & 0.618 $\uparrow$ \\ \hline
\end{tabular}
\end{table}

\section{Conclusion}
This paper presents a winning solution for the RIVA Cervical Cytology Challenge. By leveraging a powerful Co-DINO-Swin-Large baseline, coupled with center-preserving data augmentation, analytical geometric box optimization, and track-specific loss tuning, our approach achieved 1st place in Track B and 2nd place in Track A. 

Admittedly, our methodology was predominantly driven by engineering strategies to maximize the mAP evaluation metric, with relatively less exploration into the underlying clinical and biological intricacies of the cytology task itself. 
% We also want to highlight that our geometric Box Optimization technique offers a highly effective, model-agnostic approach for improving detection metrics in datasets with fixed-size annotations. We hope this engineering pipeline and our geometric insights will serve as a practical reference for researchers facing similar object detection challenges.
We also want to highlight that our geometric Box Optimization technique offers a model-agnostic refinement for detection tasks with fixed-size annotations. We hope this pipeline and our geometric insights serve as a practical reference for similar object detection challenges.

\section{ACKNOWLEDGEMENTS}
This work was supported by AI \& AI for Science Project of Nanjing University.

\bibliographystyle{IEEEbib}
\bibliography{strings,refs}

\end{document}